\documentclass{article} 
\usepackage{tencent_ailab_tech_report}
\usepackage[colorlinks = true,
            linkcolor = blue,
            urlcolor  = blue,
            citecolor = blue,
            anchorcolor = blue]{hyperref}   
\usepackage{microtype}
\usepackage{hyperref}
\usepackage{url}
\usepackage{booktabs}
\usepackage{enumitem}
\usepackage{multicol}
\usepackage{multirow}
\usepackage{CJKutf8}
\usepackage{amsmath}
\usepackage{siunitx}
\usepackage{floatflt}
\usepackage{wrapfig}
\usepackage{graphicx}
\usepackage{booktabs}
\usepackage{wrapfig}
\usepackage{authblk}
\usepackage{array}

\usepackage{lipsum}
\usepackage{makecell}

\usepackage{algorithm}
\usepackage{algorithmicx}
\usepackage{algpseudocode}
\usepackage{microtype}
\usepackage{booktabs} 
\usepackage{pifont}  
\usepackage{graphicx}  
\usepackage{subcaption} 
\usepackage{hyperref}
\usepackage{booktabs, diagbox}
\usepackage{tabu}
\usepackage{siunitx}
\newcolumntype{d}{S}          
\usepackage{caption}
\usepackage{xparse}
\usepackage{stackengine}

\usepackage{siunitx}
\usepackage{xcolor,colortbl,booktabs,tabularx,makecell,graphicx}

\usepackage{adjustbox}

\newcolumntype{R}{>{\raggedright\arraybackslash}p{3.6cm}}
\usepackage{amssymb} 

\usepackage{amsmath, amssymb}

\algnewcommand{\LeftComment}[1]{\Statex \(\triangleright\) #1}

\usepackage{array}
\usepackage{amsmath}
\usepackage{amssymb}
\usepackage{mathtools}
\usepackage{amsthm}
\usepackage{arydshln}
\usepackage[capitalize,noabbrev]{cleveref}
\usepackage{adjustbox} 
\usepackage{enumitem}
\usepackage{xr}
\usepackage{listings}
\usepackage{bbm}
\usepackage{multirow}

\usepackage[utf8]{inputenc}   
\usepackage[T1]{fontenc}       
\usepackage[english]{babel}

\theoremstyle{plain}

\theoremstyle{definition}

\theoremstyle{remark}

\usepackage[textsize=tiny]{todonotes}

\definecolor{nred}{RGB}{196, 38, 11}
\definecolor{ngreen}{RGB}{18, 141, 21}
\definecolor{nblue}{RGB}{41, 52, 190}
\definecolor{hzw}{RGB}{223, 97, 76}
\definecolor{lt}{RGB}{54, 89, 170}
\definecolor{TableGreen}{RGB}{0, 196, 0 }

\definecolor{LightGreen}{RGB}{243,247,240}

\definecolor{softgreen}{HTML}{e1f1f2}

\definecolor{soft}{HTML}{DFF0D8}

\newcommand{\ignore}[1]{}

\colmfinalcopy

\title{Every Question Has Its Own Value: \\Reinforcement Learning with Explicit Human Values}

\author[ ]{Dian Yu$^{1}$}
\author[ ]{Yulai Zhao$^{1,2}$}
\author[ ]{Kishan Panaganti$^{1}$}
\author[ ]{Linfeng Song$^{1}$} 
\author[ ]{Haitao Mi$^{1}$}
\author[ ]{Dong Yu$^{1}$}

\affil[1]{Tencent AI Lab}
\affil[2]{Princeton University}

\begin{document}

\maketitle

\begin{abstract}
We propose Reinforcement Learning with Explicit Human Values (RLEV), a method that aligns Large Language Model (LLM) optimization directly with quantifiable human value signals. While Reinforcement Learning with Verifiable Rewards (RLVR) effectively trains models in objective domains using binary correctness rewards, it overlooks that not all tasks are equally significant. RLEV extends this framework by incorporating human-defined value signals directly into the reward function. Using exam-style data with explicit ground-truth value labels, RLEV consistently outperforms correctness-only baselines across multiple RL algorithms and model scales. Crucially, RLEV policies not only improve value-weighted accuracy but also learn a value-sensitive termination policy: concise for low-value prompts, thorough for high-value ones. We demonstrate this behavior stems from value-weighted gradient amplification on end-of-sequence tokens. Ablation studies confirm the gain is causally linked to value alignment. RLEV remains robust under noisy value signals, such as difficulty-based labels, demonstrating that optimizing for an explicit utility function offers a practical path to aligning LLMs with human priorities.

\end{abstract}

\begin{figure*}[h!]
  \centering
  \includegraphics[width=0.8\textwidth]{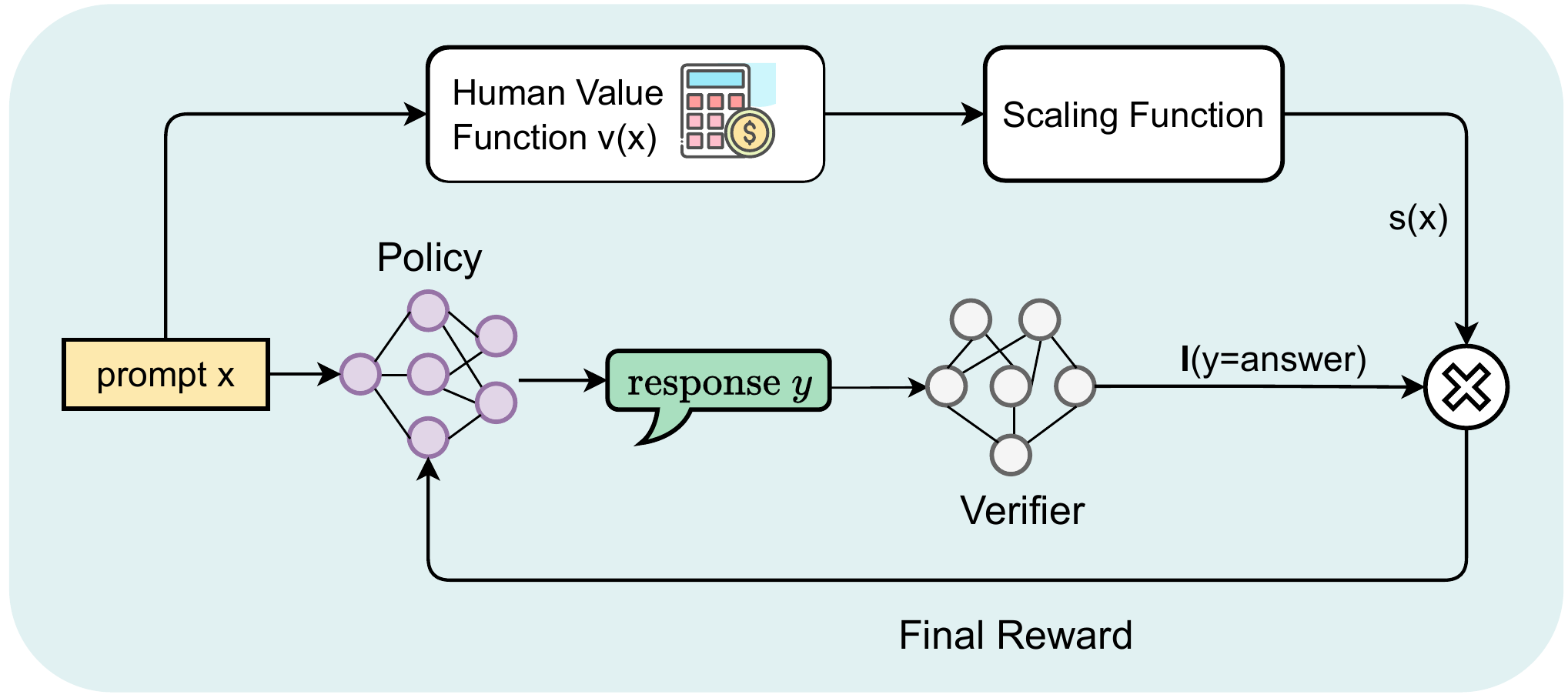}
  \caption{RLEV overview. The verifier can be either a reward model or rule-based function.}
  \label{fig:overview}
\end{figure*}

\section{Introduction}

Aligning Large Language Models (LLMs) with human goals can follow two paradigms: \textbf{implicit} value learning, which infers human utility from feedback, and \textbf{explicit} value learning, which optimizes directly for defined utility signals. The dominant paradigm, Reinforcement Learning from Human Feedback (RLHF) (\cite{stiennon2020learning,ouyang2022training,rafailov2023direct}), learns an \textbf{implicit} utility model from subjective pairwise preferences. While effective for non-verifiable tasks, this is often unnecessary for objective domains. For these, Reinforcement Learning with Verifiable Rewards (RLVR) (\cite{lambert2024t,guo2025deepseek,su2025crossing}) offers a simpler, more direct approach, using a binary reward for correctness. However, this method carries a critical oversight: by assigning a uniform reward (e.g., +1) to all correct answers, it treats all prompts as equally important, failing to capture the explicit and non-uniform value common in real-world scenarios. For instance, on an exam, correctly answering a 10-point question is demonstrably more valuable than answering a 2-point one. An LLM trained to maximize only the count of correct answers is not optimized for the total score, which is the true human objective.

To bridge this gap, we introduce \textbf{Reinforcement Learning with Explicit Human Values} (\textbf{RLEV}), a method that extends the RLVR framework by integrating explicit, human-assigned values into the reward function. RLEV operationalizes a simple principle: the utility of a response depends jointly on its correctness and the intrinsic value of its prompt. Using 100k exam-style training examples with ground-truth value labels, we show that RLEV consistently outperforms the standard RLVR baseline across multiple RL algorithms (REINFORCE++~\citep{hu2025reinforce++}, RLOO~\citep{ahmadian2024back}, and GRPO~\citep{shao2024deepseekmath}) and model scales (7B and 32B)~\citep{qwen2.5}. Notably, RLEV-trained policies learn a value-sensitive termination policy, generating highly concise responses for low-value prompts while remaining thorough on high-value ones. Our gradient analysis reveals this behavior stems from the value-scaled reward amplifying updates on end-of-sequence tokens, encouraging the model to terminate efficiently based on the prompt's importance.

Crucially, we demonstrate through ablation studies that this performance gain is causally linked to alignment with human-defined values. Baselines using randomly shuffled or uniformly scaled rewards show no significant improvement over correctness-only training. Finally, we show RLEV is robust even with noisy value signals, such as pseudo-labels from a score predictor or weak labels based on question difficulty, which still outperform the baseline. These findings establish that directly optimizing for an explicit utility function is a potent and effective method for aligning LLM behavior with stated human priorities.

Our contributions are as follows:

\begin{itemize}

\item We propose RLEV, a novel training paradigm that aligns LLMs with explicit human priorities by scaling correctness rewards with quantifiable value signals.

\item We demonstrate empirically that RLEV consistently outperforms strong correctness-only baselines across multiple RL algorithms and model scales, leading to higher value-weighted scores and a desirable property of generating more concise responses.\footnote{The RLEV dataset is available at \url{https://huggingface.co/datasets/sarosavo/RLEV}.} 

\item Through gradient analysis and ablation studies, we provide strong evidence that these gains are causally linked to value alignment, not merely to changes in reward magnitude.

\item We show that RLEV is robust and practical, achieving superior performance even when using noisy or approximate value signals, such as difficulty-based weak labels.
\end{itemize}

\section{Method: Learning from Human-Aligned Rewards}

To align a Large Language Model with human priorities, we first define a utility function that captures the desired behavior. We then operationalize this function as a scalar reward for reinforcement learning.

\subsection{A Human Utility Function for Valued and Verifiable Outcomes}

We begin from the principle that the value of a model's response depends on both its correctness and the importance of the prompt. We can formalize this by defining a \textbf{human utility function}, $U(x, y)$, for a response $y$ to a prompt $x$:
\begin{equation}
    U(x, y) = v(x) \cdot \mathbf{1}_{\text{correct}}(y)
\end{equation}
where:
\begin{itemize}
    \item $v(x)$ represents the intrinsic \textbf{human-defined value} or importance of the prompt $x$.
    \item $\mathbf{1}_{\text{correct}}(y)$ is an indicator function that is $1$ if the response $y$ is verifiably correct and $0$ otherwise.
\end{itemize}

This utility function captures the simple, powerful idea that a correct response is worth the value of the question, while an incorrect response has zero utility. The goal of our alignment process is to train a policy $\pi$ that maximizes the expected utility, $\mathbb{E}_{y \sim \pi(y|x)}[U(x, y)]$. While we instantiate this principle in exam-like settings, the same formulation applies to any domain where outcome correctness and human-assigned importance jointly determine utility, such as medical triage, tutoring, or content moderation. This product-based utility function $U(x,y)$ is a straightforward formalization of human priorities in domains where outcome correctness is verifiable and input importance is non-uniform (e.g., exams, medical triage).

\subsection{Normalizing Human Values}
\label{sec:normalization}

To obtain a practical signal for $v(x)$, we use ground-truth scores from human-designed tasks, such as exams. Since different exams have different total scores, we must normalize these values to create a consistent scale.
Let:
\begin{itemize}
    \item $s_{ij}$ be the raw score of question $j$ in exam $i$.
    \item $T_i$ be the total score of exam $i$.
\end{itemize}
We define the normalized value $v(x)$ for a given question $x$ (i.e., question $j$ in exam $i$) as its proportion of the exam total:
\begin{equation}
\label{eq:v(x)}
    v(x) = \frac{s_{ij}}{T_i}
\end{equation}
This proportional scaling naturally bounds $v(x)$ between 0 and 1 and makes it interpretable as the relative importance of the question.

\subsection{The RLEV Reward Function}
\label{sec:reward}

While $U(x,y)$ defines our target objective, its direct use as a reward can lead to unstable training. A low-value but correct question could receive a near-zero reward, discouraging the model from learning to answer it. To ensure a stable and effective learning signal, we design a practical surrogate reward function, $r(x,y)$, that preserves the relative importance of prompts while guaranteeing a minimum reward of 1 for any correct response. We achieve this by defining a scaling factor $s(x)$ based on the normalized human value $v(x)$ that is always greater than or equal to 1:

\begin{equation}
    r(x, y) = s(x) \cdot \mathbf{1}_{\text{correct}}(y)
\end{equation}
where $s(x)$ is a scaling factor based on the normalized human value $v(x)$:
\begin{equation}
    s(x) = 1 + \min(\alpha \cdot v(x), 1)
\end{equation}
Here, $\alpha$ is a scaling hyperparameter. The resulting reward $r(x, y)$ is within the range $[1, 2]$ for correct responses and is $0$ for incorrect ones. This formulation incentivizes correctness on all questions while providing a stronger ``bonus'' for correctly answering high-value ones. This additive and clipped form is chosen specifically to ensure a stable learning signal by providing a minimum reward for all correct answers while preventing excessively large rewards from destabilizing the training process, a design choice validated in our ablation studies (Section~\ref{sec:exp:design}).

\subsection{The Reinforcement Learning Objective}

We aim to find the optimal policy $\pi_{\theta}$ that maximizes the expected cumulative reward $J(\theta)$ over a dataset of prompts $\mathcal{D}$, standard in REINFORCE-style RL~\citep{williams1992simple}:
$$
J(\theta) = \mathbb{E}_{x \sim \mathcal{D}, y \sim \pi_{\theta}(\cdot|x)} \left[ \sum_{t=0}^{T-1} r(x, y_{<t}, y_t) \right],
$$
where $r(x, y_{<t}, y_t)$ denotes the per-step reward. In our setting, the reward is sparse and non-zero only at the final step $T$, thus simplifying the objective to:
$$
J(\theta) = \mathbb{E}_{x \sim \mathcal{D}, y \sim \pi_{\theta}(\cdot|x)} [r(x, y)].
$$

The corresponding gradient is:

$$
\nabla J(\theta) = \mathbb{E}_{x \sim \mathcal{D}, y \sim \pi_{\theta}(\cdot|x)} \Big[ r(x, y) \nabla \log \pi_{\theta} (y | x) \Big].
$$

Given that the policy is autoregressive, $\log \pi_{\theta} (y | x) = \sum_{t=0}^{T-1} \log \pi_{\theta} (y_t|x, y_{<t})$.

\subsection{Gradient Derivation}

To analyze the learning dynamics, we derive the policy gradients for a single prompt $x$ (noting the full gradient $\nabla J(\theta)$ is the expectation over $\mathcal{D}$) with respect to the parameters at a single time step $t$. At step $t$, we use $z_{k}$ to refer to the logit at token $k \in \mathcal{V}$ where $\mathcal{V}$ denotes the whole vocabulary. Note that $\mathcal{V}$ also includes \textsc{EOS} symbol, denoted as $e$. For any token $v \in \mathcal{V}$, we use the following to represent the conditional probability that \emph{the final $y$ is correct} given $y_t =v$, where the probability is taken over the remaining rollout under the current policy
\begin{equation}
    p_v = \Pr(\text{correct} \mid x, y_{<t}, y_t = v).
\end{equation}

Logits at step $t$ are converted to probabilities using the softmax function; then we have

\begin{equation}
\begin{aligned}
\frac{\partial}{\partial z_k} \log \pi(y_t \mid x, y_{<t})
&= \frac{\partial}{\partial z_k}
   \log \frac{\exp\!\big(z_{y_t}\big)}{\sum_{v \in \mathcal{V}} \exp(z_v)} \\
&= \mathbf{1}\{y_t = k\} - \pi(k \mid x, y_{<t}) .
\end{aligned}
\end{equation}

\begin{equation}
\begin{aligned}
\frac{\partial J}{\partial z_k}
&= \mathbb{E}_{y_t \sim \pi(\cdot \mid x, y_{<t})}
   \Big[\, r(x,y) \,\big(\mathbf{1}\{y_t = k\} - \pi(k \mid x, y_{<t})\big) \Big] \\
&= \mathbb{E}\!\big[r(x,y) \,\mathbf{1}\{y_t = k\}\big]
   - \pi(k \mid x, y_{<t})\,\mathbb{E}[r(x,y)] \\
&= \pi(k \mid x, y_{<t}) \big( \mathbb{E}[r(x,y) \mid y_t = k] - \mathbb{E}[r(x,y)] \big) 
\end{aligned}
\end{equation}
where we employ the law of total expectation in the last line.

Since $r(x,y) = s(x) \cdot \mathbf{1}_{\text{correct}}(y)$, as $s(x)$ is constant for a given $x$, we have

\begin{equation}
\begin{aligned}
\mathbb{E}[r(x,y) \mid y_t = k] = s(x) \, p_k
\end{aligned}
\end{equation}

\begin{equation}
\begin{aligned}
\mathbb{E}[r(x,y)] = s(x) \sum_{v \in \mathcal V} \pi(v \mid x, y_{<t}) p_v  
\end{aligned}
\end{equation}

Therefore, 

\begin{equation}
\begin{aligned}
\frac{\partial J}{\partial z_k}
&= \pi(k \mid x, y_{<t}) \big( \mathbb{E}[r \mid y_t = k] - \mathbb{E}[r] \big) \\
&= \pi(k \mid x, y_{<t}) \big( s(x) \, p_k -  s(x) \sum_{v \in \mathcal V} \pi(v \mid x, y_{<t}) p_v  \big) \\
&= \boxed{\pi(k \mid x, y_{<t}) s(x) \cdot \big( p_k - \sum_{v \in \mathcal V} \pi(v \mid x, y_{<t}) p_v  \big)} \\
\end{aligned}
\end{equation}

Consider the special \textsc{EOS} token $e$, which is also in $\mathcal{V}$. Below, we investigate the training dynamics of this special token. First we define ${\overline{p}}_{\neg e}$ as the averaged probability of final correctness over all non-\textsc{EOS} tokens:

\begin{equation}
{\overline{p}}_{\neg e} := \frac{1}{1- \pi_e} \sum_{v \neq e, v \in \mathcal{V}} \pi_v p_v.
\end{equation}

Then the advantage for choosing \textsc{EOS} is

\begin{equation}
\label{eq:e}
\mathbb{E}[r \mid y_t=e] - \mathbb{E}[r]
= s(x)(1-\pi_e)\big(p_e - \overline{p}_{\neg e}\big).
\end{equation}

Thus the gradient with respect to the \textsc{EOS} logit is:

\begin{equation}
\label{eq:g}
\boxed{
\frac{\partial J}{\partial z_e}
= s(x) \cdot \pi_{e}(1-\pi_e)\big(p_e - \overline{p}_{\neg e}\big).}
\end{equation}

The resulting gradient for the \textsc{EOS} logit (Equation~\ref{eq:g}) is driven by the difference between the expected correctness of terminating the sequence ($p_e$) and the average expected correctness of continuing ($\overline{p}_{\neg e}$), scaled by the human value factor $s$.

\begin{itemize}
    \item \textsc{EOS} receives a positive gradient if its correctness probability exceeds the average continuation correctness, i.e.\ $p_e > \overline{p}_{\neg e}$.
    \item A continuation token $c$ receives a negative gradient if $p_c < \sum_{v} \pi_v p_v$ (the policy-weighted average). Thus, when $p_e > \overline{p}_{\neg e}$, most continuations are pushed down, though any $c$ with $p_c$ above the global average can still receive a positive update.
    \item The human-aligned scale $s \in [1,2]$ multiplies the gradient magnitude without changing these conditions. Therefore, when $p_e > \overline{p}_{\neg e}$ (i.e.\ there is already sufficient information for correctness), \textsc{EOS} is reinforced more strongly, which accelerates the tendency to end earlier.    
\end{itemize}

In summary, compared to a purely binary correctness reward, this scheme encourages the policy to generate more concise, more accurate completions. Moreover, because the reward is scaled by the human-defined scoring function rather than correctness alone, the resulting policy is expected to achieve higher human-defined scores in real-world use, which is supported by our experimental results in Section~\ref{sec:experiment}. \textbf{Multiplying by the human-value factor amplifies the gradient's magnitude, which more strongly reinforces the decision to terminate when correctness is already likely}.

\section{Experiments}
\label{sec:experiment}

\subsection{Datasets}

The dataset comprises question-answering pairs from multi-subject exams, with the original content predominantly in Chinese. The reference answers are written by domain experts for objective human evaluation, making them suitable for RLVR. Additionally, we extract each question's human-labeled score and the total score of the exam it originates from.   Subsequently, we partition the data into training and testing sets containing 100,000 and 8,000 examples, respectively. We split by exam to avoid leakage.

To assess the generalization ability of the RLEV policies trained on Chinese data, we evaluate the out-of-distribution performance on several English and Chinese general-domain benchmarks (GPQA Diamond~\citep{rein2024gpqa}, C-Eval~\citep{huang2023c}, MMLU-Pro~\citep{wang2024mmlu}, and SuperGPQA~\citep{du2025supergpqa}).

As ground-truth human-defined values may be unavailable in many scenarios, in Section~\ref{sec:exp:rlev_with_other_types}, we investigate the effectiveness of RLEV with two types of ``noisy'' human values. We conduct experiments using WebInstruct-verified~\citep{general-reasoner}, a general domain English dataset with objective answers. We map each of the five difficulty category (\textsc{Primary School}, \textsc{Junior High School}, \textsc{Senior High School}, \textsc{University}, and \textsc{PhD}) into value scores (1, 2, 4, 6, 8), respectively. We divide the score by 100 for normalization. For each category, we randomly sample 2,000 training examples and train with the resulting 10k instances.

To make this resource more accessible to the broader research community, we used GPT-4o to translate the data (with human-labeled values) into English, which will also be released.

\subsection{Experimental Setup}
We kept the training setup consistent for all estimators. All policies were trained for one epoch on eight GPUs with a learning rate of 5e-7. The rollout batch size was set to 128. The maximum length for both prompts and generated responses was capped at 1024 tokens. For evaluation, we use greedy decoding. We use base models (Qwen2.5-7B and Qwen2.5-32B~\citep{qwen2.5}) for policy initialization.

\subsection{Evaluation Metrics}
To evaluate our method, we use a set of metrics designed to capture both correctness and alignment with human-defined values:

\paragraph{Accuracy (Acc):} The standard, unweighted accuracy calculated as the percentage of total correct responses. This metric measures correctness without considering the value of each prompt.

\paragraph{Human-Aligned Accuracy (H-Acc):} The value-weighted accuracy, calculated as the ratio of achieved value from correct responses to the total possible value:
$$
\text{H-Acc} = \frac{\sum_{\text{correct responses}} v(x)}{\sum_{\text{all responses}} v(x)}
$$
    
\paragraph{Response Length (Resp. Length):} The average number of tokens in a model's generated response.
    
\paragraph{Value Density:} An efficiency metric measuring value delivered per token, calculated by dividing the \textbf{H-Acc value expressed as a percentage} by the average response length. This is particularly relevant for tasks focused on verifiable correctness, where the primary goal is to provide the correct answer efficiently.

Following previous RLVR studies for general domains~\citep{su2025crossing,general-reasoner}, we use a large language model (Qwen2.5-72B-Instruct~\citep{qwen2.5}) to verify the semantic equivalence between the final answer of a response and the reference answer. This automated verification method has been widely shown to have high agreement with human annotators in objective, non-adversarial, reference-based evaluation settings~\citep{zhao2025one}. Importantly, focusing verification on only the final part of the response did not cause length collapse in our experiments.

\subsection{RLEV with Ground-Truth Human Values}

Our primary results show that RLEV consistently outperforms the correctness-only baseline across all tested configurations. This holds true for both 7B and 32B models, which see average Human-Aligned Accuracy (H-Acc) gains of 2.0\% and 2.8\%, respectively (Table~\ref{tab:merged_results}). This improvement is driven by a learned focus on high-value tasks; as detailed in the appendix (Table~\ref{tab:merged_results:value_bin}), the accuracy gains are generally notably larger for high-valued prompts than for low-valued ones.

A key benefit is a value-sensitive termination policy, which will be discussed in Section~\ref{sec:concise}. The model learns to be concise on low-value prompts while remaining thorough on high-value ones. This leads to an overall increase in conciseness. For example,  RLEV models more than halve the average response length, from 246.9 to 98.6 tokens for the 32B models.

This efficiency and strategic improvement also generalize effectively. Even though trained on Chinese data, the RLEV models outperform their correctness-only counterparts on several out-of-distribution (OOD) English and Chinese benchmarks, with the 32B model showing notable gains on tasks like GPQA Diamond and SuperGPQA (Table~\ref{tab:ood}).

\begin{table}[h!]
\centering
\caption{Comparison of policies trained with RLEV (human-aligned) and baseline (correctness) rewards across 7B and 32B models. RLEV consistently improves accuracy and conciseness for both model scales.}
\label{tab:merged_results}
\begin{tabular}{lllcccc}
\toprule
\textbf{Estimator} & \textbf{Size} & \textbf{Reward Type} & \textbf{Acc} & \textbf{H-Acc} & \textbf{Resp. Length}  & \textbf{Value Density}\\
\midrule
\multirow{4}{*}{REINFORCE++} & \multirow{2}{*}{7B}  & correctness    & 63.8           & 55.0           & 168.1 & 0.33 \\
                             &                      & \cellcolor{softgreen}human-aligned & \cellcolor{softgreen}\textbf{65.3} & \cellcolor{softgreen}\textbf{57.0} & \cellcolor{softgreen}84.8 &\cellcolor{softgreen}\textbf{0.67}  \\
\cmidrule(lr){2-7}
                             & \multirow{2}{*}{32B} & correctness    & 67.7           & 57.6           & 226.2 & 0.25 \\
                             &                      & \cellcolor{softgreen}human-aligned & \cellcolor{softgreen}\textbf{71.0} & \cellcolor{softgreen}\textbf{61.9} & \cellcolor{softgreen} 68.7 & \cellcolor{softgreen}\textbf{0.90}  \\
\midrule
\multirow{4}{*}{RLOO}        & \multirow{2}{*}{7B}  & correctness    & 65.9           & 56.7           & 186.2 & 0.30\\
                             &                      & \cellcolor{softgreen}human-aligned & \cellcolor{softgreen}\textbf{66.6} & \cellcolor{softgreen}\textbf{58.9} & \cellcolor{softgreen}86.4 &\cellcolor{softgreen}\textbf{0.68} \\
\cmidrule(lr){2-7}
                             & \multirow{2}{*}{32B} & correctness    & 70.9           & 60.9           & 345.5 & 0.18\\
                             &                      & \cellcolor{softgreen}human-aligned & \cellcolor{softgreen}\textbf{72.3} & \cellcolor{softgreen}\textbf{63.3} & \cellcolor{softgreen} 78.7 & \cellcolor{softgreen}\textbf{0.80}  \\
\midrule
\multirow{4}{*}{GRPO}        & \multirow{2}{*}{7B}  & correctness    & 65.7           & 56.0           & 251.1 & 0.22\\
                             &                      & \cellcolor{softgreen}human-aligned & \cellcolor{softgreen}\textbf{66.2} & \cellcolor{softgreen}\textbf{57.7} & \cellcolor{softgreen}100.4 &\cellcolor{softgreen}\textbf{0.57} \\
\cmidrule(lr){2-7}
                             & \multirow{2}{*}{32B} & correctness    & 70.6           & 59.9           & 169.0 & 0.35 \\
                             &                      & \cellcolor{softgreen}human-aligned & \cellcolor{softgreen}\textbf{71.3} & \cellcolor{softgreen}\textbf{61.7} & \cellcolor{softgreen}148.3 & \cellcolor{softgreen}\textbf{0.42}\\
\midrule
\multirow{4}{*}{Average}     & \multirow{2}{*}{7B}  & correctness    & 65.1           & 55.9           & 201.8 & 0.28\\
                             &                      & \cellcolor{softgreen}human-aligned & \cellcolor{softgreen}\textbf{66.0} & \cellcolor{softgreen}\textbf{57.9} & \cellcolor{softgreen}90.5 &\cellcolor{softgreen}\textbf{0.64} \\
\cmidrule(lr){2-7}
                             & \multirow{2}{*}{32B} & correctness    & 69.7           & 59.5           & 246.9 & 0.26 \\
                             &                      & \cellcolor{softgreen}human-aligned & \cellcolor{softgreen}\textbf{71.5} & \cellcolor{softgreen}\textbf{62.3} & \cellcolor{softgreen}98.6  & \cellcolor{softgreen}\textbf{0.71}\\
\bottomrule
\end{tabular}
\end{table}

\begin{table}[h!]
\centering
\caption{OOD Results across English and Chinese general-domain tasks.}
\label{tab:ood}
\begin{tabular}{lcccc}
\toprule
\textbf{Model} & \textbf{GPQA Diamond } & \textbf{C-Eval} & \textbf{MMLU-Pro} & \textbf{SuperGPQA} \\
\midrule
Base-7B &  \textbf{31.8} & 60.8 & 45.0 & 25.4 \\
$~~~$+ correctness & \textbf{31.8} & 76.2 & 51.5 & 26.2  \\
\rowcolor{softgreen}$~~~$+ human-aligned & 31.3 & \textbf{76.4} & \textbf{52.5}  & \textbf{26.8}  \\

\midrule
Base-32B &33.2  &  57.9  & 55.1 & 33.2\\
$~~~$+ correctness & 39.9 & 84.9 & \textbf{63.0} &  34.0 \\
\rowcolor{softgreen}$~~~$+ human-aligned & \textbf{43.4} & \textbf{85.4} & \textbf{63.0} &  \textbf{36.2} \\
\bottomrule
\end{tabular}
\end{table}

\subsection{RLEV with Other Types of Human Values}
\label{sec:exp:rlev_with_other_types}

\begin{table}[h!]
\centering
\caption{RLEV performance with imperfect value signals on the test set of WebInstruct-verified. We test two ``noisy'' value sources: weak labels derived from task difficulty and predictor-generated values from a score predictor trained on our main exam dataset. Both methods consistently outperform the correctness-only baseline, showing RLEV's robustness when ground-truth values are unavailable.}
\label{tab:weak_labels}
\begin{tabular}{lcccccc}
\toprule
\textbf{Model} & \textbf{Acc | H-Acc} & \textbf{primary} & \textbf{junior} & \textbf{senior} & \textbf{university} & \textbf{phd}\\
\midrule
Base-7B & 18.8 | 17.0  & 38.9  & 28.4 &24.5 &  13.9 & 0.0 \\
\midrule
REINFORCE++ \\
 $~~~$+ correctness  &  19.1 | 16.9 & 50.0  & 27.0   &  27.3 & 12.7 & 0.0 \\
\rowcolor{softgreen}$~~~$+ weak-labeled values & 21.2 | 19.3 &  38.9 & 28.4  & 29.4 & 15.3 & 0.0 \\
\rowcolor{softgreen}$~~~$+ predicted values & \textbf{21.6} | \textbf{19.6}  & 50.0  & 32.4  &  27.9 & 15.8 &  10.0\\

\midrule
RLOO \\
$~~~$+ correctness &   20.0 | 18.0  & 38.9 & 28.4  & 28.2 & 13.9 & 0.0 \\
\rowcolor{softgreen}$~~~$+ weak-labeled values &  20.3 | 18.4  & 33.3    &  29.7  & 28.5 &  14.1 &  10.0 \\\rowcolor{softgreen}$~~~$+ predicted values &  \textbf{21.3} | \textbf{19.1} & 38.9  & 32.4  & 30.0  & 14.6 &  0.0\\
\midrule

GRPO \\
$~~~$+ correctness &   19.4 | 17.0 & 44.4  & 33.8   & 27.0  & 12.8 & 0.0\\
\rowcolor{softgreen}$~~~$+ weak-labeled values & \textbf{20.6} | \textbf{18.7}  &  50.0  &  31.1 & 25.5 & 15.8  &  0.0\\
\rowcolor{softgreen}$~~~$+ predicted values & 20.3 | 18.2 & 50.0  &  25.7  & 28.8 & 14.1 &  0.0\\
\bottomrule
\end{tabular}
\end{table}

This result (Table~\ref{tab:weak_labels}) demonstrates RLEV's robustness and practicality. It shows the method is effective even when precise, ground-truth scores are unavailable, making it applicable to a much wider range of real-world scenarios where only heuristic value estimates (like task priority or difficulty) exist. Note that primary and phd only have 18 and 10 instances, respectively, while the total test set has 1,000 instances. We use the multi-subject exam data for training a score predictor for generating the predicted values in Table~\ref{tab:weak_labels}. We discuss the training details in Appendix~\ref{appendix:prediction}.

\subsection{Analysis of Value-Sensitive Termination}
\label{sec:concise}

\begin{figure}[ht!]
    \centering
    
    \subfloat[Baseline (Top 200 valued prompts)]{%
        \centering
        \includegraphics[width=0.48\textwidth]{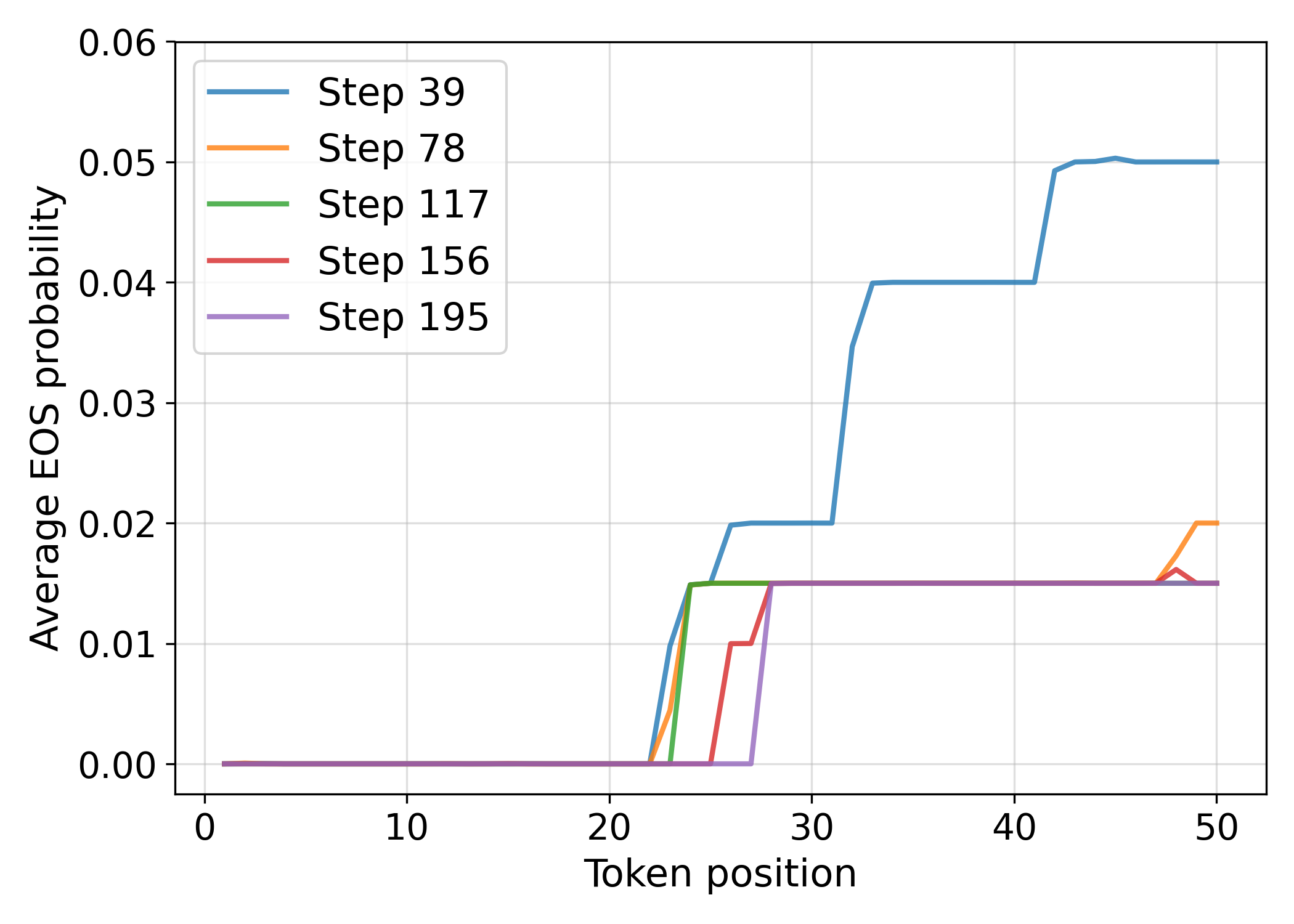}
        \label{fig:sub1}
    }
    \hfill 
    \subfloat[RLEV (Top 200 valued prompts)]{%
        \centering
        \includegraphics[width=0.48\textwidth]{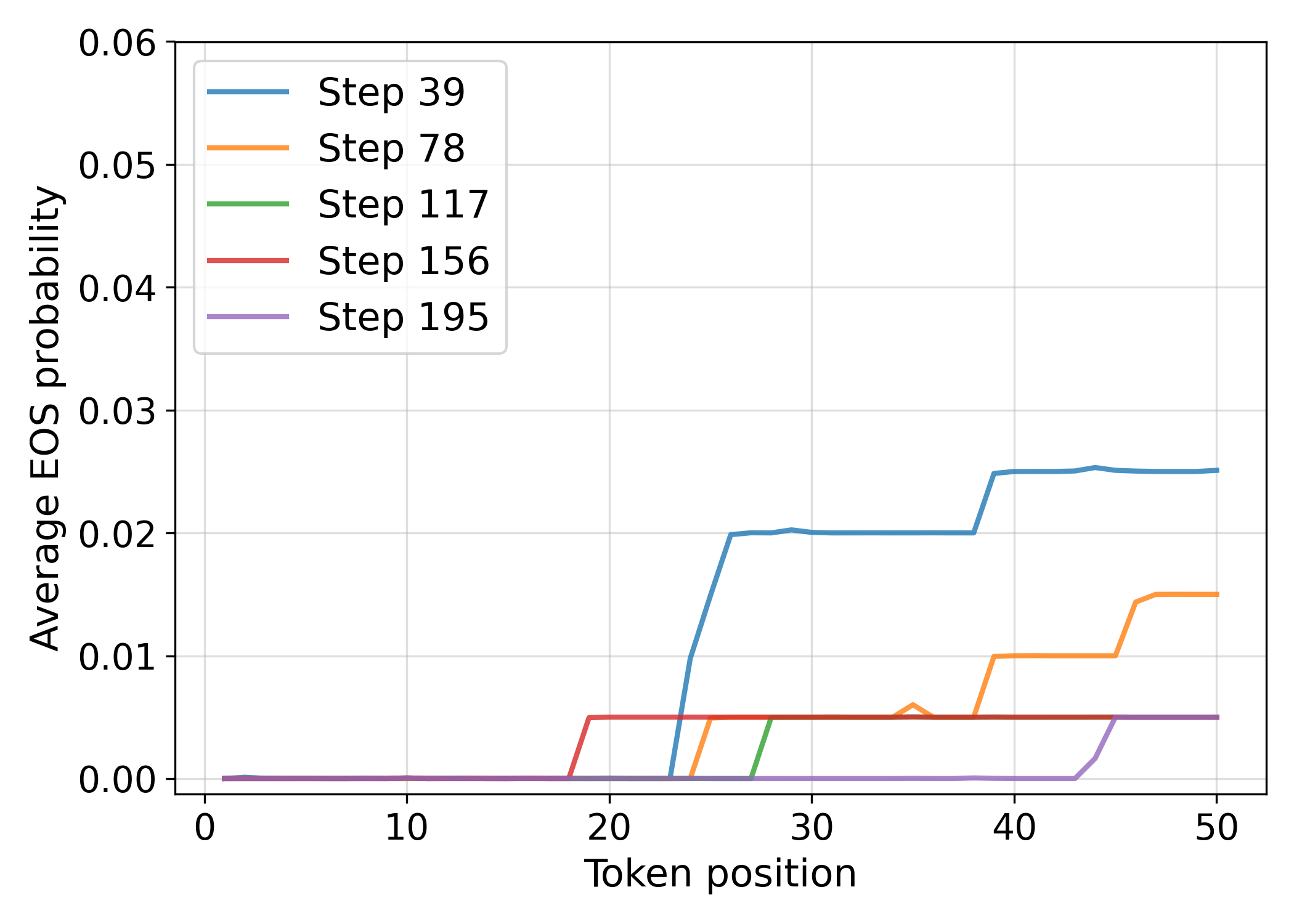}
        \label{fig:sub2}
    }
    
    \subfloat[Baseline (Bottom 200 valued prompts)]{%
        \centering
        \includegraphics[width=0.48\textwidth]{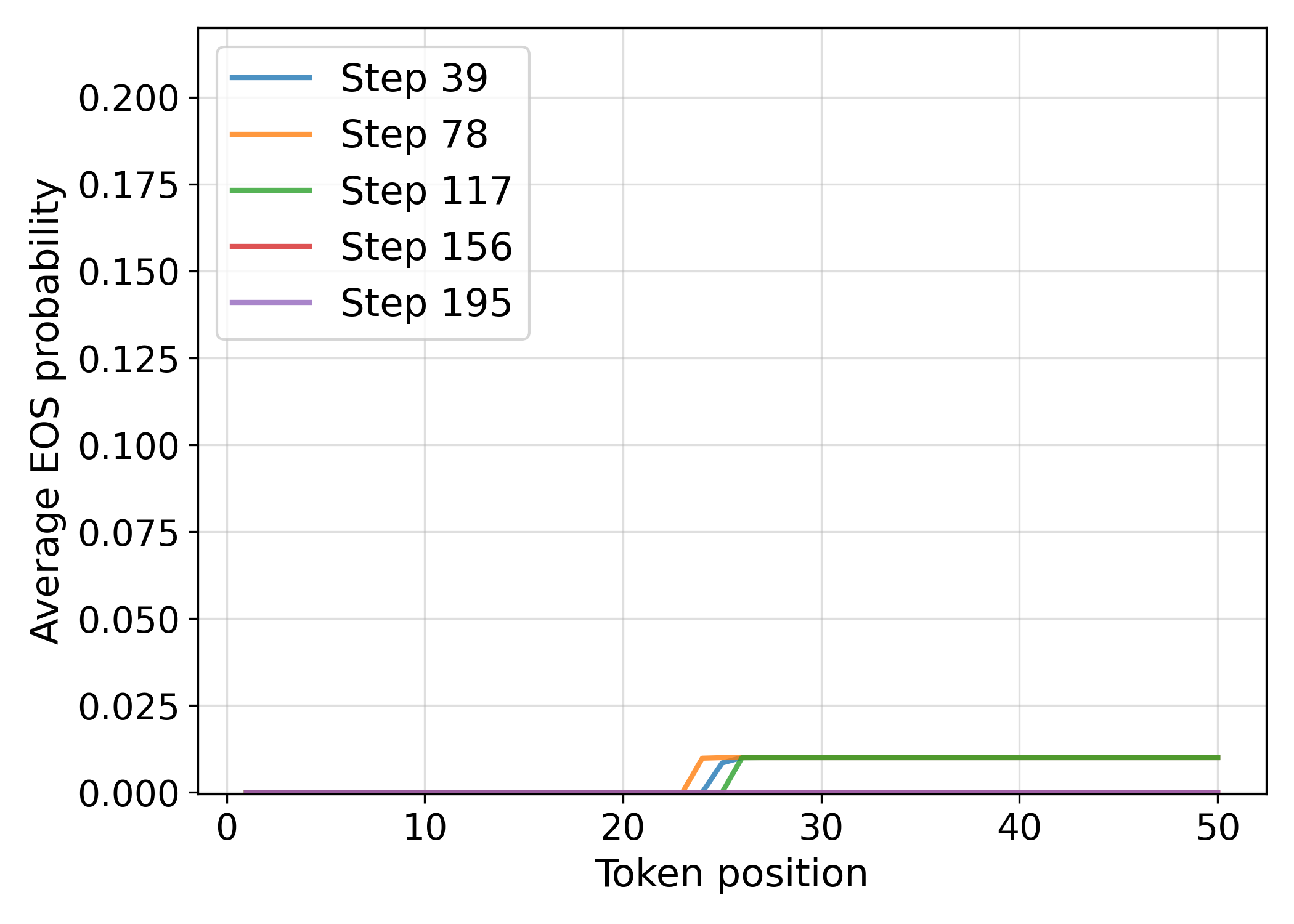}
        \label{fig:sub3}
    }
    \hfill 
    \subfloat[RLEV (Bottom 200 valued prompts)]{%
        \centering
        \includegraphics[width=0.48\textwidth]{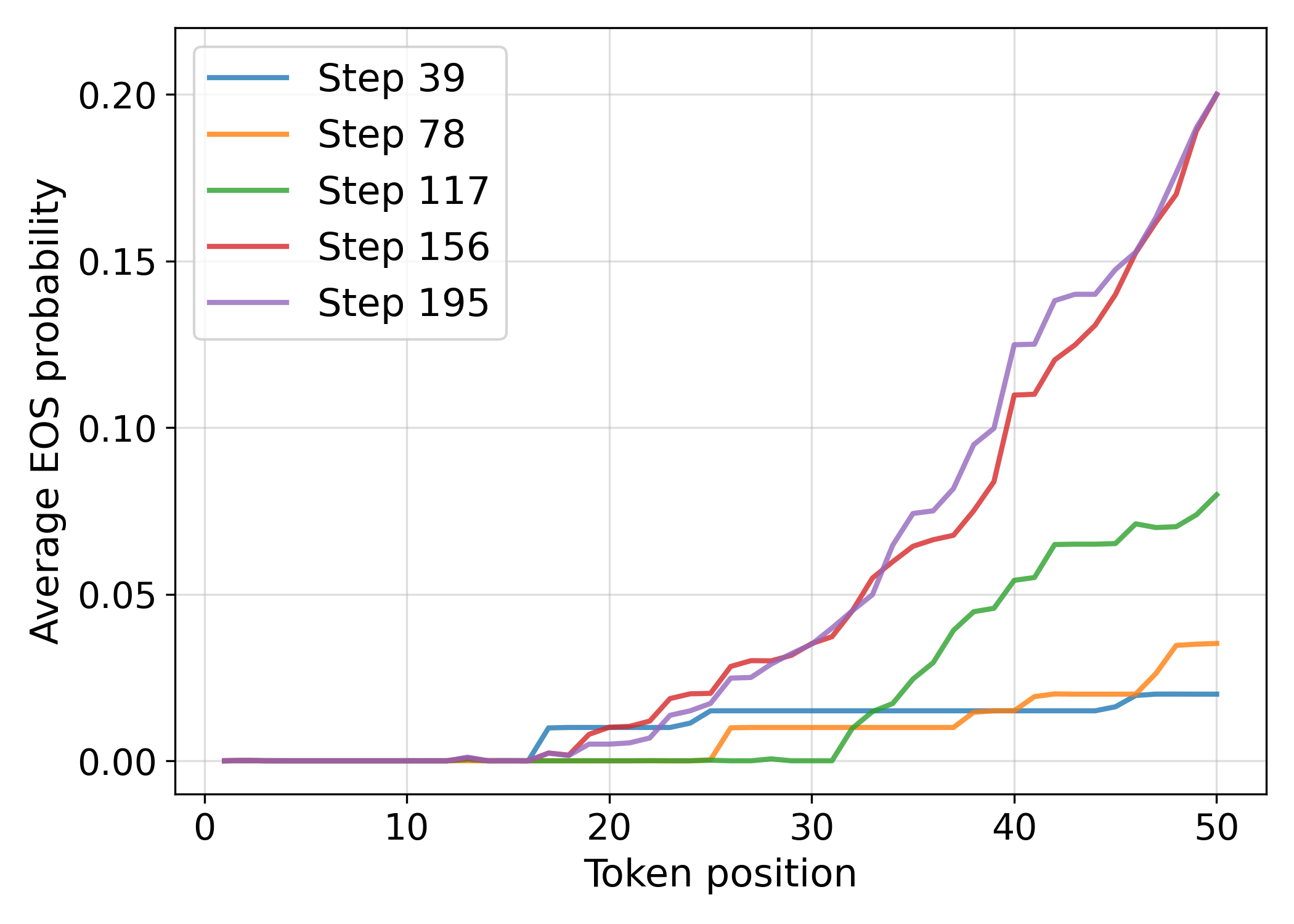}
        \label{fig:sub4}%
    }
    
    \caption{EOS probability trajectories for RLEV and the baseline, showing different termination policies for high-value (top) and low-value (bottom) prompts.}
    \label{fig:four_figures}
\end{figure}

Token-level Analysis: The hypothesis from the gradient analysis (``value-scaling amplifies updates on the EOS token'') is directly validated by our token-level analysis, though the behavior is more nuanced than a simple uniform increase in EOS probability. As shown in Figure~\ref{fig:four_figures}, the RLEV model learns a sophisticated, value-sensitive termination policy

For \textbf{low-value} prompts, the RLEV model assigns a dramatically higher probability to the EOS token much earlier in the generation process compared to the baseline. Once a sufficient answer is generated for these simpler prompts, the value-weighted reward strongly reinforces the decision to stop, leading to highly concise outputs.

Conversely, for \textbf{high-value} prompts, which are often more complex, the RLEV model learns to suppress the probability of the EOS token relative to the baseline. This behavior encourages the model to generate a more thorough and complete response. The gradient analysis explains this as the large value-scaling factor $s$ amplifying the signal to continue when the expected correctness from adding more tokens is higher than from stopping prematurely.

This dual mechanism shows that RLEV does not merely learn to be shorter; it learns to allocate its token budget strategically, being efficient on low-stakes questions while being cautious and comprehensive on high-stakes ones to maximize the overall human-aligned score.

\subsection{Ablation Studies: Isolating the Impact of Human Values}

A key claim of our work is that aligning the reward signal with human-defined values is responsible for the observed performance gains. However, an alternative hypothesis is that the improvements stem from simply increasing the magnitude of the rewards for correct answers, rather than the value-alignment itself. To isolate the effect of human-aligned values, we conduct several ablation studies. Besides the correctness-only baseline, we also compare our full RLEV (human-aligned) model against two controls:

\paragraph{Uniform Scaling:} All correct responses receive a constant $\bar{s}$, where $\bar{s}$ is the average reward scale calculated across the training prompts ($\bar{s} = \mathbb{E}[s(x)] \approx 1.2$) (details in Appendix~\ref{appendix:data_statistics}). This control is designed to test the alternative hypothesis that \textbf{general increase in reward magnitude, irrespective of its alignment with prompt value, is sufficient to cause the observed performance gains}.

\paragraph{Random Weights:} The reward is scaled using the RLEV formula, but the human values $v(x)$ are randomly shuffled across the training set before being used to calculate the scaling factor $s(x)$. This procedure creates a placebo reward signal that maintains the exact same distribution of reward magnitudes as the primary experiment but completely decouples the reward from the prompt's true value. This directly tests whether the \textbf{causal factor for improvement is the specific alignment} between higher rewards and higher-value prompts.

\begin{table}[h!]
\centering
\caption{Ablation study results using the RLOO estimator. Uniformly scaling the reward degrades performance, while using random weights does not improve conciseness. Only when the reward scaling is directly correlated with human-defined values do we see a meaningful increase in human-aligned accuracy (h-acc) and a desirable reduction in response length.}
\label{tab:ablation}
\begin{tabular}{@{}lcccc@{}}
\toprule
\textbf{Reward Scaling Method} & \textbf{Acc} & \textbf{H-Acc} & \textbf{Resp. Length} & \textbf{Value Density}\\
\midrule
correctness (baseline) & 65.9 & 56.7 & 186.2 & 0.30 \\
uniform scaling             & 65.3 & 55.1 & 358.4 &  0.15 \\
random weights (shuffled)   & 66.4 & 57.4 &  280.5 & 0.20 \\
\textbf{human-aligned (ours)} & \textbf{66.6} & \textbf{58.9} & 86.4 & \textbf{0.68} \\
\bottomrule
\end{tabular}
\end{table}

\subsection{Reward Function Sensitivity and Design}
\label{sec:exp:design}

To validate our reward function design, we analyze its sensitivity to both the hyperparameter \(\alpha\) and its specific mathematical form.

\paragraph{Sensitivity to Hyperparameter \(\alpha\)}
The choice of \(\alpha\) is crucial as it determines how strongly the human value \(v(x)\) influences the final reward. We trained models using our primary reward function, \(r(x,y) = (1 + \min(\alpha \cdot v(x), 1)) \cdot \mathbf{1}_\text{correct}(y)\), with several values of \(\alpha\). As shown in Table~\ref{tab:alpha_sensitivity}, while performance is robust across a range of values, we found that \(\alpha=10\) offered the best balance of human-aligned accuracy and response conciseness.

\begin{table}[h]
\centering
\caption{Sensitivity to hyperparameter \(\alpha\). Performance is reported across all key metrics.}
\label{tab:alpha_sensitivity}
\begin{tabular}{@{}lcccc@{}}
\toprule
\textbf{Hyperparameter \(\alpha\)} &  \textbf{Acc} & \textbf{H-Acc} & \textbf{Resp. Length}& \textbf{Value Density} \\
\midrule
baseline & 65.9 & 56.7 & 186.2 & 0.30 \\
\midrule
1  & 66.4 &  58.1 & 101.5 & 0.57 \\
5   & 66.1  &  56.8 & 141.0 & 0.40 \\
\textbf{10}  & \textbf{66.6} & \textbf{58.9} & 86.4 & 0.68 \\
15 & 66.3 & 58.1 &  62.4  & \textbf{0.93} \\
20 & 66.1 & 56.8 & 157.9 & 0.36\\
\bottomrule
\end{tabular}
\end{table}

\paragraph{Ablation on Reward Function Form}

To justify our choice of an additive and clipped reward scaler, we compare it against a purely multiplicative alternative: \(r(x,y) = (1 + \alpha \cdot v(x)) \cdot \mathbf{1}_\text{correct}(y)\). Table~\ref{tab:reward_form_ablation} shows that our chosen form yields superior results.

There are two possible reasons: first, the mean $v(x)$ is $0.02$, and only $1.18\%$ of the training examples have a value $> 0.1$. This highly right-skewed distribution, which is visualized in Appendix~\ref{appendix:data_statistics} (Figure~\ref{fig:histogram:train}), indicates that for over $98\%$ of the data, our function acts as a fine-grained linear reward scaler, preserving the original human value. Second, for the small fraction of high-value outliers shown in the distribution's tail, the clipping mechanism prevents the excessively large rewards that the purely multiplicative form would generate, thus stabilizing training process and leading to better overall performance.

\begin{table}[h]
\centering
\caption{Comparison of different reward scaling functions.}
\label{tab:reward_form_ablation}
\begin{tabular}{@{}lcccc@{}}
\toprule
\textbf{Reward Function Form} & \textbf{Acc } & \textbf{H-Acc } & \textbf{Resp. Length}   & \textbf{Value Density}\\
\midrule
purely multiplicative & 66.4 & 57.6 & 201.6 & 0.29 \\
\textbf{additive \& clipped (ours)} & \textbf{66.6} & \textbf{58.9} & 86.4 & \textbf{0.68} \\
\bottomrule
\end{tabular}
\end{table}

\section{Related Work}

The idea of weighting learning signals according to their relative importance has deep roots in classical RL. Early methods such as importance-weighted transfer~\citep{tirinzoni2018importance}, reward-weighted regression~\citep{peters2007reinforcement}, and advantage-weighted regression~\citep{peng2019advantage} all adjust gradient updates to emphasize high-value samples. These studies show that non-uniform weighting can improve sample efficiency or align behavior with desired utility, but they do not consider the case where each data point (e.g., a prompt or question) carries a human-defined point value reflecting its real-world importance.

Recent work in the LLM alignment domain has focused on RL with Verifiable Rewards~\citep{luong2024reft,lambert2024t,guo2025deepseek,su2025crossing}, which train models using objective correctness signals. Other studies have proposed shaping or enriching verifiable rewards: for example, ConfClip~\citep{zhang2025confclip}, rubrics as rewards~\citep{gunjal2025rubrics}, and composite reward frameworks such as RLCR with calibration rewards~\citep{damani2025beyond}. While these approaches modify reward form or composition, they do not explicitly scale correctness rewards by human-assigned per-prompt values normalized across a dataset, nor analyze the resulting gradient-level mechanisms.

Our method, RLEV, integrates human-assigned per-prompt importance into the RLVR framework using a clipped scaling surrogate. Through empirical tests, ablations, and gradient analysis, RLEV yields more concise and human-aligned behavior by optimizing for explicit, value-weighted utility, enabling alignment with explicitly defined human utility functions.

\section{Conclusions and Future Work}

We introduced Reinforcement Learning with Explicit Human Values (RLEV), a paradigm that aligns LLMs with human priorities by scaling correctness rewards with an explicit value signal. Experiments show RLEV consistently outperforms correctness-only baselines, improving value-weighted accuracy and leading to the generation of more concise responses. We trace this conciseness to value-weighted gradient amplification on end-of-sequence (EOS) tokens. Ablation studies confirm these gains are causally linked to value alignment rather than reward magnitude. Furthermore, the method proves robust, surpassing baselines even with noisy value signals derived from task difficulty.

Future work could explore more dynamic value functions that are learned or adapt to user priorities. Another promising direction is to combine RLEV, for grounding in objective correctness and importance, with RLHF to fine-tune subjective qualities like style and tone. This hybrid approach could offer a more holistic path to LLM alignment.

\section{Broader Impact and Limitations}
Ultimately, this work demonstrates that directly optimizing for an explicit, non-uniform utility function is a robust and effective method for aligning LLM behavior with human priorities. By moving beyond simple binary rewards, RLEV encourages models to develop a more nuanced understanding of value, learning not just what constitutes a correct answer, but also how much each correct answer matters. This represents a practical step toward creating systems that are not only more capable but also more judicious in applying their capabilities to what humans deem most important. Despite its effectiveness, this work has several limitations. First, the framework formulates human value as a single, pre-defined scalar quantity suited for objective domains where importance is explicitly quantified. However, human values in a broader sense are often complex, multi-dimensional, and subjective. Second, applying RLEV requires explicit value labels for each prompt. While our experiments show RLEV is robust to noisy signals, this data dependency remains a practical consideration. Finally, the current method relies on a static value function, and future work could explore more dynamic value functions that adapt to user priorities in real-time.

\bibliography{neurips_2024}
\bibliographystyle{colm2024_conference}

\clearpage
\newpage
\appendix
\section{Appendix}

\subsection{Data Statistics}
\label{appendix:data_statistics}
\begin{figure*}[h!]
  \centering
  \includegraphics[width=0.9\textwidth]{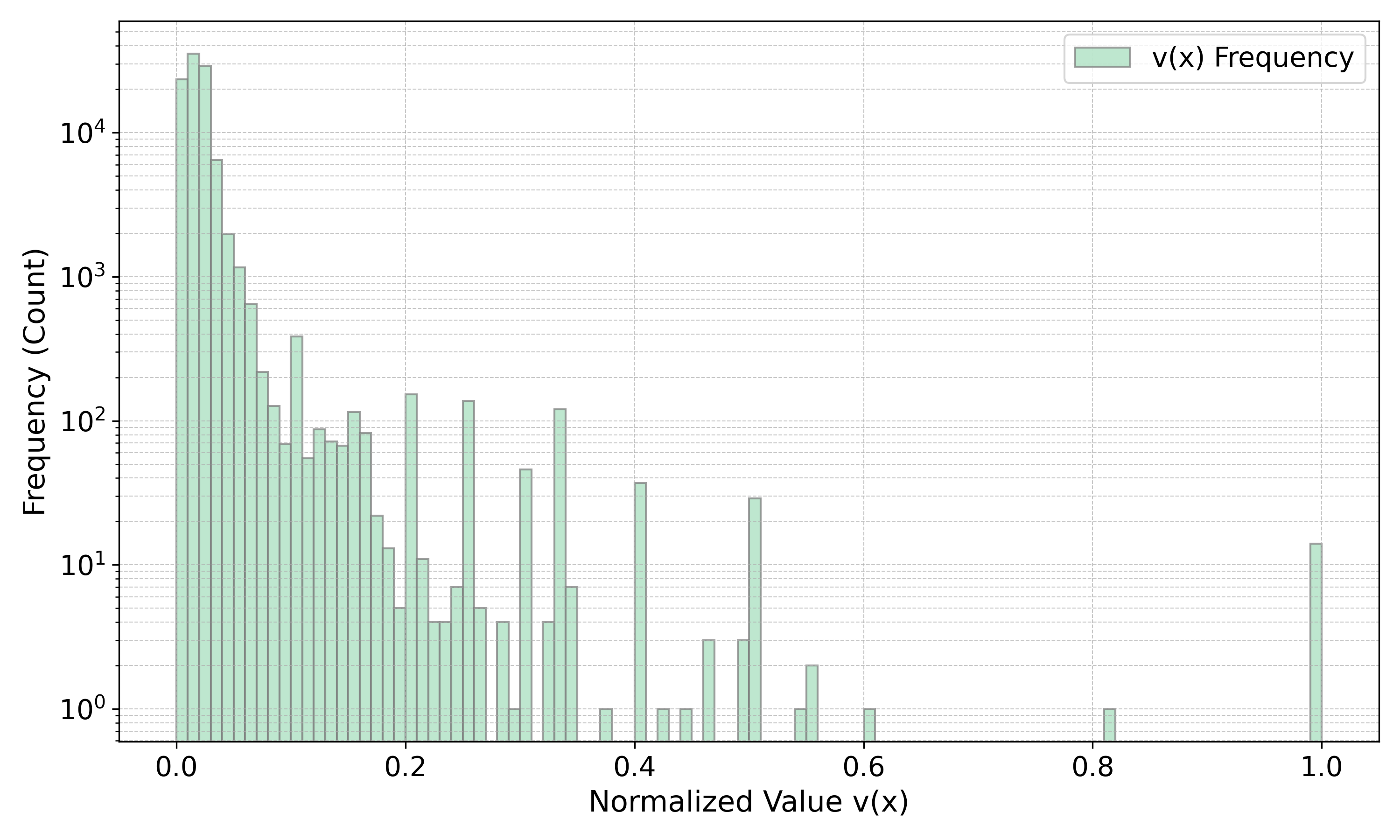}
  \caption{Distribution of human-defined normalized values v(x) in training data (100k) with ground-truth values.}
  \label{fig:histogram:train}
\end{figure*}

\begin{figure*}[h!]
  \centering
  \includegraphics[width=0.9\textwidth]{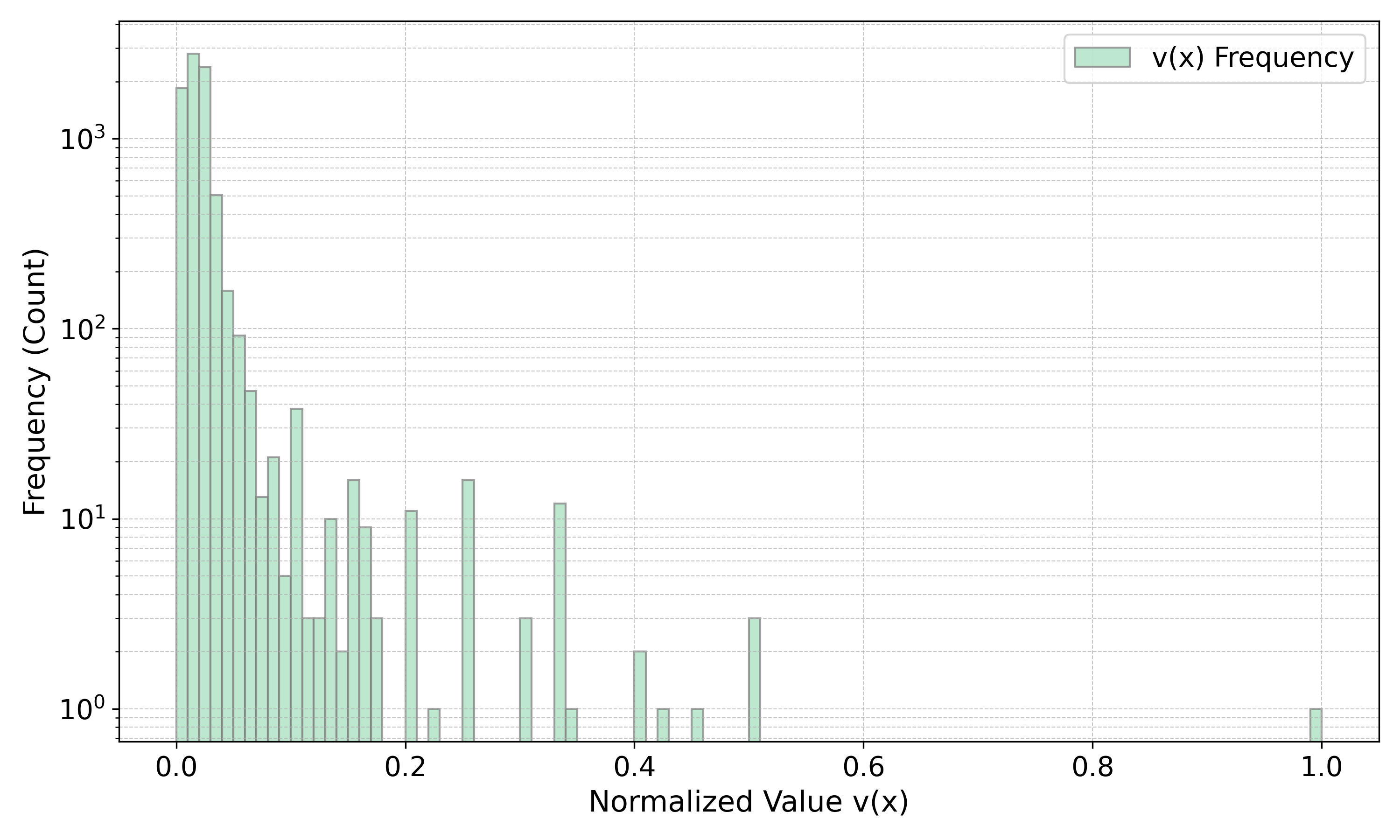}
  \caption{Distribution of human-defined normalized values v(x) in test data (8k) with ground-truth values.}
  \label{fig:histogram:test}
\end{figure*}

We analyze the human-defined values in 100k training instances and the 8k testing instances. These scores are normalized per-exam proportional scores defined in Equation~\ref{eq:v(x)} (Section~\ref{sec:normalization}). See distribution of normalized values in the training and test subsets in Figure~\ref{fig:histogram:train} and Figure~\ref{fig:histogram:test}.

\subsection{Score Prediction}
\label{appendix:prediction}

\begin{table}[h!]
\centering
\caption{Prompt Structure for the Score Predictor}
\label{tab:prompt_structure}
\begin{tabular}{@{}ll@{}}
\toprule
\textbf{Role} & \textbf{Content} \\ \midrule
\texttt{system} & \parbox[t]{0.8\linewidth}{
\texttt{f``You are a scoring assistant. Given a question and its answer, output a numeric score greater than 0 and up to \{total\_score\} inclusive (decimals allowed, e.g. 0.5) that reflects how much this problem would contribute in a \{total\_score\}-point exam. Respond with only the score, no other text.''}
} \\ \\
\texttt{user}   & \texttt{question} \\ \bottomrule
\end{tabular}
\end{table}

To evaluate RLEV's performance with imperfect value signals, besides rule-based scores derived from difficulty levels, we train a score predictor to generate pseudo values for datasets where ground-truth scores are unavailable.

We convert the exam data into the format shown in Table~\ref{tab:prompt_structure} and train the score predictor with supervised fine-tuning for two epochs using Qwen2.5-7B. For datasets such as WebInstruct-verified, we standardize the task by setting a consistent total score of 100 for all prompts. We use the same test set for evaluating the performance of the score predictor. It achieves an exact-match accuracy of 79.5\%. The Pearson correlation between the predicted and true scores is 0.91 (p < 0.001), indicating a strong positive relationship.

\subsection{Detailed Accuracy Analysis}

\begin{table}[h!]
\centering
\caption{Comparison of policies trained with RLEV (human-aligned) and baseline (correctness) rewards across 7B and 32B models. We also report the accuracy on top 20\% high-valued prompts and bottom 20\% low-valued prompts.}
\label{tab:merged_results:value_bin}
\begin{tabular}{lllcccc}
\toprule
\textbf{Estimator} & \textbf{Size} & \textbf{Reward Type} & \textbf{Acc (all)} & \textbf{Acc (high-valued)} & \textbf{Acc (low-valued)}  \\
\midrule
\multirow{4}{*}{REINFORCE++} & \multirow{2}{*}{7B}  & correctness    & 63.8           & 54.5          & 68.9  \\
                             &                      & \cellcolor{softgreen}human-aligned & \cellcolor{softgreen}\textbf{65.3} & \cellcolor{softgreen}\textbf{58.0} & \cellcolor{softgreen}\textbf{69.8}  \\
\cmidrule(lr){2-7}
                             & \multirow{2}{*}{32B} & correctness    & 67.7                    & 57.6 & 73.4 \\
                             &                      & \cellcolor{softgreen}human-aligned & \cellcolor{softgreen}\textbf{71.0}  &\cellcolor{softgreen}\textbf{62.9} & \cellcolor{softgreen}\textbf{76.2}  \\
\midrule
\multirow{4}{*}{RLOO}        & \multirow{2}{*}{7B}  & correctness    & 65.9           & 57.4           & 71.6\\
                             &                      & \cellcolor{softgreen}human-aligned & \cellcolor{softgreen}\textbf{66.6} & \cellcolor{softgreen}\textbf{58.8} & \cellcolor{softgreen}\textbf{71.8}  \\
\cmidrule(lr){2-7}
                             & \multirow{2}{*}{32B} & correctness    & 70.9           & 60.9           & 76.6\\
                             &                      & \cellcolor{softgreen}human-aligned & \cellcolor{softgreen}\textbf{72.3} & \cellcolor{softgreen}\textbf{62.1} & \cellcolor{softgreen}\textbf{78.1}  \\
\midrule
\multirow{4}{*}{GRPO}        & \multirow{2}{*}{7B}  & correctness    & 65.7           & 55.7           & 71.9\\
                             &                      & \cellcolor{softgreen}human-aligned & \cellcolor{softgreen}\textbf{66.2} & \cellcolor{softgreen}\textbf{57.1} & \cellcolor{softgreen}\textbf{72.4}  \\
\cmidrule(lr){2-7}
                             & \multirow{2}{*}{32B} & correctness    & 70.6           &  59.3          &  76.8 &  \\
                             &                      & \cellcolor{softgreen}human-aligned & \cellcolor{softgreen}\textbf{71.3} & \cellcolor{softgreen}\textbf{61.0} & \cellcolor{softgreen}76.6 \\

\bottomrule
\end{tabular}
\end{table}

As shown in Table~\ref{tab:merged_results:value_bin}, human-aligned (RLEV) policy achieves a higher accuracy than the correctness baseline in all high-valued bins and nearly all low-valued bins. The improvement is generally more obvious for the high-valued prompts. These results show that RLEV specifically guides the model to perform better on the prompts that are defined as more valuable or important.

\end{document}